%% file: main.tex
\documentclass[10pt,twocolumn,letterpaper]{article}

\usepackage[pagenumbers]{cvpr} %

\usepackage{graphicx}
\usepackage{amsmath}
\usepackage{amssymb}
\usepackage{booktabs}
\usepackage{color, colortbl}
\definecolor{LightCyan}{rgb}{0.88,1,1}
\definecolor{Gray}{gray}{0.9}

\usepackage[pagebackref,breaklinks,colorlinks]{hyperref}

\usepackage[capitalize]{cleveref}
\crefname{section}{Sec.}{Secs.}
\Crefname{section}{Section}{Sections}
\Crefname{table}{Table}{Tables}
\crefname{table}{Tab.}{Tabs.}

\def\cvprPaperID{5938} %
\def\confName{CVPR}
\def\confYear{2022}

\input{preamble}

\begin{document}

\input{sec/0_metadata}

\twocolumn[{
\renewcommand\twocolumn[1][]{#1}%
\maketitle
\input{fig/teaser}
}]

\input{sec/0_abstract}
\input{sec/1_introduction}
\input{sec/2_related}

\input{sec/3_method}
\input{sec/4_dataset}

\input{sec/5_experiments}
\input{sec/6_conclusions}
{
    \small
    \bibliographystyle{ieee_fullname}
    \bibliography{macros,main}
}

\input{sec/X_supplementary}

\end{document}

%% file: preamble.tex
\usepackage{overpic}
\usepackage{enumitem} %
\usepackage{overpic} %
\usepackage{color}
\usepackage{tabularx}
\definecolor{turquoise}{cmyk}{0.65,0,0.1,0.3}
\definecolor{purple}{rgb}{0.65,0,0.65}
\definecolor{dark_green}{rgb}{0, 0.5, 0}
\definecolor{orange}{rgb}{0.8, 0.6, 0.2}
\definecolor{red}{rgb}{0.8, 0.2, 0.2}
\definecolor{darkred}{rgb}{0.6, 0.1, 0.05}
\definecolor{blueish}{rgb}{0.3, 0.3, .6}
\definecolor{light_gray}{rgb}{0.7, 0.7, .7}
\definecolor{pink}{rgb}{1, 0, 1}
\definecolor{greyblue}{rgb}{0.25, 0.25, 1}
\definecolor{awesome}{rgb}{1.0, 0.13, 0.32}

\usepackage{blindtext}

\renewcommand{\paragraph}[1]{\vspace{1em}\noindent\textbf{#1}.}

\usepackage{gensymb}

\usepackage{graphicx}               %

%% file: sec/0_metadata.tex
\title{Revealing Occlusions with 4D Neural Fields}

\author{Basile Van Hoorick$^1$\hspace{.2cm} Purva Tendulkar$^1$\hspace{.2cm} D\'idac Sur\'is$^1$\hspace{.2cm} Dennis Park$^2$\hspace{.2cm} Simon Stent$^2$\hspace{.2cm} Carl Vondrick$^1$\\[0.09cm]
$^1$Columbia University\hspace{.4cm} $^2$Toyota Research Institute\\[0.01cm]
{\tt\footnotesize \{basile, purvaten, didacsuris, vondrick\}@cs.columbia.edu, \{dennis.park, simon.stent\}@tri.global}
}

%% file: fig/teaser.tex
\begin{center}
\vspace{-0.7cm}
\centering
\captionsetup{type=figure}
\includegraphics[width=\linewidth]{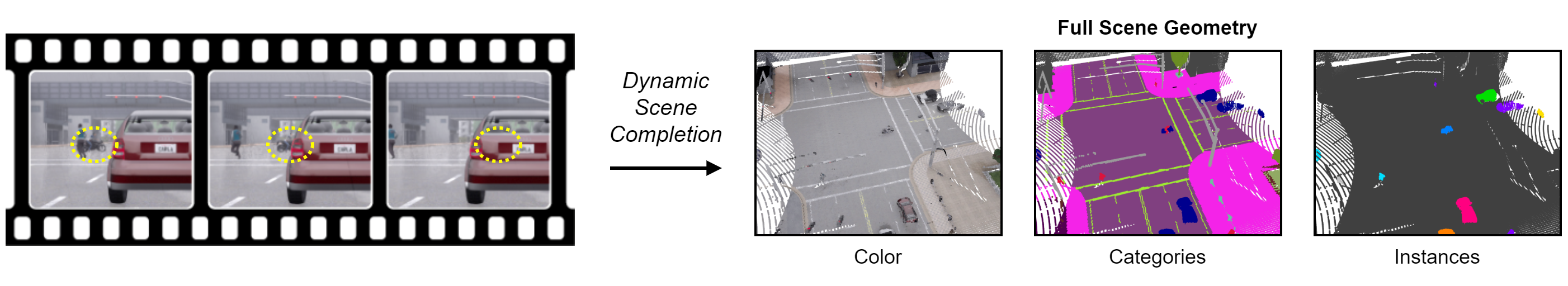}
\vspace{-0.6cm}
\captionof{figure}{
\textbf{Video Occlusions --}
Although the motorcycle (circled in yellow) becomes fully occluded in the video, we can still perform many visual recognition tasks, such as predicting its location, reconstructing its appearance, and classifying its semantic category. This paper introduces a video representation architecture that is able to learn to perform all of these occlusion reasoning tasks. We show example inputs and ground truths of the proposed dynamic scene completion framework.
\vspace{0.1cm}
}
\label{fig:teaser}
\end{center}%

%% file: sec/0_abstract.tex
\begin{abstract}
\vspace{-0.2cm}
For computer vision systems to operate in dynamic situations, they need to be able to represent and reason about object permanence. We introduce a framework for learning to estimate 4D visual representations from monocular \mbox{RGB-D} video, which is able to persist objects,
even once they become obstructed by occlusions. 
Unlike traditional video representations, we encode point clouds into a continuous representation, which permits the model to attend across the spatiotemporal context to resolve occlusions.
On two large video datasets that we release along with this paper, our experiments show that the representation is able to successfully reveal occlusions for several tasks, without any architectural changes. 
Visualizations show that the attention mechanism automatically learns to follow occluded objects.
Since our approach can be trained end-to-end and is easily adaptable, we believe it will be useful for handling occlusions in many video understanding tasks. Data, code, and models are available at \url{occlusions.cs.columbia.edu}.
\vspace{-0.3cm}
\end{abstract}

%% file: sec/1_introduction.tex
\section{Introduction}
\label{sec:intro}

When an object becomes occluded in video, its location and visual structure is often still predictable. In several studies, developmental psychologists have been able to demonstrate that shortly after birth, children learn how objects persist during occlusions \cite{spelke1993perceiving,baillargeon1985object,piaget2013construction,aguiar19992}, and evidence suggests that animals perform similar reasoning too \cite{pepperberg1990object,marino2017thinking}.\footnote{See ``\href{https://www.youtube.com/results?search_query=what+the+fluff+challenge}{What The Fluff Challenge}’’ on YouTube.} For example, although the yellow orb in Figure \ref{fig:teaser} disappears behind other objects, its location, geometry, and appearance remain evident to you. Occlusions are fundamental to computer vision, and predicting the contents behind them underlies many applications in video analysis.

The field has developed a number of deep learning methods to operate on point clouds \cite{qi2017pointnet,yuan2018pcn,zhao2021point,guo2020pct,yang2019pointflow}, which due to their attractive properties, have emerged as the representation of choice for numerous 3D tasks. Point clouds are sparse, making them particularly scalable to large scenes. However, to solve the problem in Figure \ref{fig:teaser}, we need a video representation that (1) uses evidence from the previous frames in order to (2) generate the new points that are not observed in the subsequent frames. Since point clouds are possible to collect at scale \cite{chang2019argoverse,caesar2020nuscenes}, we believe they are an excellent source of data for learning to predict behind occlusions in video. However, the representation must also have the capacity to \emph{create} points conditioned on their context.

This paper introduces an architecture for learning to predict 4D point clouds from an RGB-D video camera. The key to our approach is a continuous neural field representation of a point cloud, which uses an attention mechanism to condition the full space on the observations. Since the representation is continuous, the approach can learn to produce points anywhere in spacetime, allowing for high-fidelity reconstructions of complex scenes. Where there are occlusions and missing observations, the representation is able to use attention to find the object and missing scene structure when it was last visible, and subsequently put the right points in the right place.

Experiments show that our video representation learns to successfully perform many occlusion reasoning tasks, such as visual reconstruction, geometry estimation, tracking, and semantic segmentation. The same method works for these tasks without architectural changes. On two different datasets, we show the approach remains robust for both highly cluttered scenes and objects of various sizes. Though we train the representation without ground truth correspondence, visualizations show that the attention mechanism automatically learns to follow objects through occlusions.

There are three principal contributions in this paper. Firstly, we propose the new fundamental task of \textbf{4D dynamic scene completion}, which forms a basis for spatiotemporal reasoning tasks. Secondly, we present new benchmarks to evaluate scene completion and object permanence in cluttered situations. Thirdly, we introduce a new architecture for deep learning on point clouds, which is able to generate new points conditioned on their context. This architecture allows for large-scale point cloud data to be leveraged for representation learning. In the remainder of the paper, we describe these contributions in detail. We invite the community to use these benchmarks to test their model's video understanding capabilities.

%% file: sec/2_related.tex
\section{Related Work}
\label{sec:related}

Learning to persist objects through occlusions has been a long-standing challenge in computer vision~\cite{nguyen2001occlusion,huang2005tracking,pan2007robust}.
In recent years, researchers have combined modern deep learned features with a variety of approaches to track through occlusions.
These include classical Kalman filtering or linear extrapolation~\cite{khurana2021detecting}, 2D recurrent neural networks~\cite{tokmakov2021learning}, and more explicit reasoning mechanisms~\cite{shamsian2020learning}.
Our approach tackles the problem in a more holistic manner, drawing on improvements in point cloud modeling, neural fields, and attention mechanisms. We briefly recap relevant work from each area.

\textbf{Point cloud modeling.}
Earlier work on representing point clouds with deep networks is based on 2D projection~\cite{su2015mvc, cvpr17chen, kanezaki2021_rotationnet} or 3D voxelization~\cite{Maturana2015voxnet, song2016ssc}.
These methods capitalize on the success of 2D and 3D convolutions in image and video understanding by preprocessing input point clouds into 2D or 3D grids.
PointNet~\cite{qi2017pointnet} proposed to use point-wise MLPs and pooling layers to compute permutation-invariant point cloud representations.
PointNet was subsequently extended to allow for hierarchical features to better model local geometric structures~\cite{qi2017pointnetplusplus}, and combined with the idea of voxelization to create a highly efficient point cloud encoder~\cite{lang2019pointpillar}.
More recently, researchers have begun to apply transformer attention mechanisms that were first found to be valuable in the language domain~\cite{vaswani2017attention} to encode point clouds~\cite{Xie2018CVPR, liu2019point2sequence, lee19set}. 
To address the quadratic complexity in attention computation applied to large-scale point clouds, the Point Transformer~\cite{zhao2021point} replaced global attention with local vector attention and introduced relative position encoding. 
We adopt the Point Transformer as our feature encoder backbone because of its efficiency and performance for various point cloud tasks.

\textbf{Point cloud tasks.}
Our goal is somewhat similar to that of point completion networks~\cite{yuan2018pcn, tchapmi2019topnet, huang2020pf, wen2020point, wang2020cascaded}, although these works typically operate on a per-object basis and address only \emph{self-}occlusions or amodal completion. In contrast, we aim to reconstruct entire scenes and address the fundamental challenge of occlusions more generally.
Because existing 4D architectures~\cite{choy20194d, thomas2019kpconv, liu2019meteornet} lack a mechanism to efficiently create new points, they have not been demonstrated to be capable of dynamic scene completion.
For example, in 4D panoptic LiDAR segmentation~\cite{aygun20214dpanoptic}, the goal is to jointly tackle semantic and instance segmentation in 3D space over time. While our work addresses related tasks, we wish to be able to model not just the visible, but also the \emph{occluded} parts of the scene, by drawing on past observations or priors. This is especially valuable when spatial inputs are sparse, as they often are in LiDAR applications.

\textbf{Neural fields.}
Neural implicit functions have become very popular for 3D representation in recent years~\cite{chibane2020neural,sitzmann2020implicit,rist2021semantic,yu2020pixelnerf}, building on the seminal work of Neural Radiance Fields (NeRF)~\cite{mildenhall2020nerf} and neural implicit surface modeling \cite{park2019deepsdf}.
The basic idea of NeRF is to learn to represent a scene using a fully connected deep neural network, whose inputs are a 3D point and viewing direction and whose outputs are an estimated color and volume density.
This is attractive because it avoids the need to discretize space and can encode a scene more efficiently and richly than traditional representations such as meshes or voxels, which themselves can be extracted from the implicit model.
Numerous efforts have been made to extend NeRF to dynamic scenes~\cite{pumarola2021dnerf,park2021nerfies,gao2021freeviewvideo,du2021nerflow,xian2021space}, but in addition to requiring per-scene retraining, occlusions are typically explicitly ignored by applying losses over the non-occluded scene only.
\textbf{Transformers in vision.}
The attention mechanism introduced in~\cite{bahdanau2014neural,vaswani2017attention} has been applied with great success to computer vision~\cite{dosovitskiy2020vit, steiner2021augreg, chen2021outperform, xu2020learning}. Recently, architectures that are built solely with self-attention as computational units have started to perform on par with or better than convolutional networks as generic feature extractors~\cite{pan2021scalable} in standard vision tasks such as object detection and segmentation~\cite{liu2021swin,Ranftl2021, Ranftl2020} and point cloud--based detection~\cite{misra2021-3detr}. The role of cross-attention has also been extended as a mechanism for sensor fusion. DETR3D~\cite{detr3d} extends DETR~\cite{detr} by computing the keys and values from multi-view images. Recently, Perceiver \cite{jaegle2021perceiver} showed that asymmetric attention mechanisms can distill inputs from multiple modalities (\ie vision or point clouds) into robust latent representations.

%% file: sec/3_method.tex
\section{Approach}
We introduce the new task of 4D dynamic scene completion from posed monocular RGB-D video input.
Let $\mathcal{X} = \{\, (\boldsymbol p_i, t_i, \boldsymbol x_i) \,\}$ be a point cloud video captured from a single camera view.\footnote{A \emph{point cloud video} assumes known camera parameters to deproject the RGB + depth information into some canonical coordinate system.}
Each discrete point $(\boldsymbol p_i, t_i, \boldsymbol x_i)$ has a spatial position $\boldsymbol p_i \in \mathbb{R}^3$, a time $t_i \in \mathbb{R}$, and an RGB color $\boldsymbol x_i \in \mathbb{R}^3 $ where the subscript $i$ indicates the index.
This information can be obtained realistically using a regular camera coupled with either a depth camera or a LiDAR sensor, aggregating data over multiple frames.
Note that the input point cloud is only a partial scan, and consequently there are missing points due to occlusions or other noise, which makes this a challenging task.
Our goal is to learn a mapping from $\mathcal{X}$ to a complete point cloud 
$\mathcal{Y} = \{\, (\boldsymbol p_j, t_j, \boldsymbol y_j) \,\}$ that densely encodes the full spacetime volume. 
The output vector $\boldsymbol y_j \in \mathbb{R}^d$ encodes any labels that we want to predict, such as color or semantic category.

\input{fig/arch}

\subsection{Model}

Point clouds are often treated as discrete, which causes them to have an irregular structure that makes traditional deep representation learning on them difficult. In order for our model to learn to persist points after they become occluded, we need a mechanism to create new points that have not been observed.

We will model the output point cloud as continuous, which allows us to compactly parameterize all the putative points across the 4D spacetime volume. Let $(\boldsymbol p_q, t_q) \in \mathbb{R}^4$ be a continuous spacetime query coordinate. Our model estimates the features $\hat{\boldsymbol y}$ located at $(\boldsymbol p_q, t_q)$, which may be occluded, with the decomposition:
\begin{align}
    \hat{\boldsymbol y}(\boldsymbol p_q, t_q) = f(\boldsymbol p_q, t_q; \phi(\mathcal X))
\end{align}
where $\phi$ is a feature extractor and $f$ is our continuous representation. There are many possible choices for $\phi(\mathcal{X})$~\cite{qi2017pointnet,qi2017pointnetplusplus,yuan2018pcn}, and we use the architecture from the Point Transformer network~\cite{zhao2021point}, which produces contextualized features for every point in the (subsampled) input.

The model is able to continuously predict a representation  $\hat{\boldsymbol y}$ for the entire spacetime volume, shown in Figure \ref{fig:arch}.
We can train $\hat{\boldsymbol y}$ for many different point cloud tasks, providing us the flexibility to predict, for example, geometry, semantic information, color, or object identity.

Our model uses a continuous representation similar to methods in neural rendering and computer graphics \cite{pumarola2021dnerf,park2021nerfies,gao2021freeviewvideo,du2021nerflow,xian2021space}, which also enjoy significant computational advantages from the compact scene representation. 
However, our approach operates on point clouds instead of signed distance functions or radiance fields, which allows us to train and apply our model for many tasks besides view synthesis. 
Furthermore, our approach is conditioned on a set of frames in a dynamic point cloud video, enabling the model to learn a rich spatiotemporal representation for occluded objects.

\input{fig/keyidea}

\subsection{Point Attention}

Given the query coordinate $(\boldsymbol p_q, t_q)$, we need to estimate the contents at that spatiotemporal location. However, in a video with occlusions, the contextual evidence for those contents might be both spatially and temporally far away. 

We introduce a cross-attention layer that uses the query coordinates to attend to the input video in order to generate this prediction. We illustrate this process in Figure \ref{fig:keyidea}. Typically, attention works by using a query to attend to keys and retrieve the relevant values. In our case, we will operate on the featurized point cloud $\mathcal{Z} = \phi(\mathcal X) = \{\, (\boldsymbol p_i, \boldsymbol \alpha_i) \,\}$. We form the keys $\boldsymbol K$ and values $\boldsymbol V$ from $\boldsymbol \alpha_i$, and the relative positional encodings $\boldsymbol \Delta$ from $\boldsymbol p_q - \boldsymbol p_i$. We form the query $\boldsymbol Q$ from $\boldsymbol p_q$ and $t_q$.

Our layer can be recursively stacked, allowing the model to build increasingly rich representations of the scene.
We implement the above vector cross-attention strategy through the following calculations, inspired by \cite{zhao2021point}:
\begin{align}
    \boldsymbol Q_q &= \boldsymbol w_{Q}^\intercal \boldsymbol \beta_{q,n} & \textrm{(queries)} \\
    \boldsymbol V_i &= \boldsymbol w_{V}^\intercal \boldsymbol \alpha_i & \textrm{(values)} \\
    \boldsymbol K_i &=  \boldsymbol w_{K}^\intercal \boldsymbol \alpha_i & \textrm{(keys)} \\
    \boldsymbol \Delta_{q,i} &= \boldsymbol w_{\Delta}^\intercal (\boldsymbol p_i - \boldsymbol p_q)  & \textrm{(positions)}
\end{align}
where $\boldsymbol \beta_{q,n}$ is a feature vector that encodes the features for query point $(\boldsymbol p_q, t_q)$. The base case is $\boldsymbol\beta_{q,0} = \textrm{MLP}(\boldsymbol p_q, t_q)$, and as we stack the cross-attention block, it will be iteratively refined with:
{\small
\begin{align}
    \boldsymbol \beta_{q,n+1} = \sum_{i \in \mathcal{M}(q)} \rho( \gamma(\boldsymbol Q_q - \boldsymbol K_i + \boldsymbol \Delta_{q,i})) \odot (\boldsymbol V_i + \boldsymbol \Delta_{q,i})
    \label{eqn:ca}
\end{align}
}%
where
$\mathcal{M}(q)$ is a set of nearest neighbors within $\phi(\mathcal{X})$ around $\boldsymbol p_q$,
$\rho$ is the softmax operation for normalization,
$\gamma$ is a mapping MLP that produces the attention weights,
and $\odot$ is an element-wise product to represent per-channel feature modulation \cite{zhao2021point}. %

We apply the operation in Equation (\ref{eqn:ca}) twice, meaning we terminate the recursion at $\boldsymbol\beta_{q,2}$. This produces a feature vector that describes the contents at the query location, which we decode into the predicted labels. Finally, we use an MLP to map $\boldsymbol\beta_{q,2}$ to $\hat{\boldsymbol y}(\boldsymbol p_q, t_q)$.

\input{fig/mvsuper}

\subsection{Learning and Supervision}

We train the model for 4D dynamic scene completion.
Given several camera views of a scene, we assume known camera parameters to deproject their recordings into point clouds.
We select one camera view to be the input view, which creates $\mathcal{X}$. To form the target $\mathcal{Y}$, we use the point cloud that merges all the camera views together. We train the model to predict the multi-view point cloud $\mathcal{Y}$ from the single-view point cloud $\mathcal{X}$, illustrated in Figure \ref{fig:supervis}.

Due to the efficiency of our representation, we can train the model end-to-end for large spacetime volumes on standard GPU hardware.
We minimize the loss function:
\begin{align}
    \min_{f, \phi} \; \mathbb{E}_{(\mathcal X, \mathcal Y)} \left[\sum_{(\boldsymbol p_q, t_q, \boldsymbol y_q) \in \mathcal Y \cup \mathcal N} \mathcal{L}\left(\hat{\boldsymbol y}(\boldsymbol p_q, t_q), \boldsymbol y_q\right) \right] 
\end{align}
where $\mathcal{N}$ is a set of negative points randomly sampled uniformly from $\mathbb{R}^4$. Since the training data $\mathcal{Y}$ only contains solid points, the negative points cause the model to learn to distinguish which regions are empty space.

\subsection{Tasks}

Our framework is able to learn to reveal occlusions for several different tasks on point clouds. For every query point, the model produces a vector $\hat{\boldsymbol y}_i \in \mathbb{R}^d$, and we can supervise different dimensions of $\hat{\boldsymbol y}_i $ for various tasks. We select the loss function $\mathcal{L}$ depending on the dataset and task. We describe several options for the loss terms below.

\textbf{Geometry completion} distinguishes solid objects ($\sigma=1$) from free space ($\sigma=0$) within the scene, where the ground truth occupancy $\sigma$ is inferred for every query point by thresholding its proximity to the target point cloud. Denoting $\hat y_\sigma$ as the relevant dimension of the $\hat{\boldsymbol y}$ vector, %
we apply a standard binary cross-entropy comparison as follows: $\mathcal{L}_\sigma = \mathcal{L}_{BCE}(\hat y_\sigma, \sigma)$.

\textbf{Visual reconstruction} means that, in addition to completing the missing regions, the model must also predict a color $\hat{\boldsymbol y}_c$ in RGB space.
For the loss function, we use the $\mathcal L_1$-distance between the relevant output dimensions and the target $\boldsymbol c$: $\mathcal{L}_{\boldsymbol c} = \|\hat y_{\boldsymbol c} - \boldsymbol c\|_1$.
    
\textbf{Semantic segmentation} classifies every query point into $S$ possible categories. The output is supervised with a cross-entropy loss between the predicted categories $\hat{\boldsymbol y}_s$ and the ground truth semantic label $s$: $\mathcal{L}_{\boldsymbol s} = \mathcal{L}_{CE}(\hat{\boldsymbol y}_s, s)$

\textbf{Instance tracking} tasks the model with localizing an object, even through total occlusions, that was highlighted with a mask in only the first frame.\footnote{This is similar to most \emph{semi-supervised video object segmentation} setups \cite{caelles20182018, wang2019fast}, but in 3D space instead. Note that the object may be partially, but not completely, occluded at the beginning of the video for this to work.}
To do this, we add an extra dimension $\tau_i$ to the input point cloud $\mathcal{X}$, that indicates which points belong to the object of interest. We then train the model to propagate this indicator throughout the rest of the video, where $\hat{y}_\tau$ is the relevant dimension in the output. We use the binary cross-entropy loss between the tracking flag $\hat y_\tau$ and $\tau$: $\mathcal{L}_\tau = \mathcal{L}_{BCE}(\hat y_\tau, \tau)$.

These four loss terms can be linearly combined to form the overall objective:
\begin{align}
    \mathcal{L} &= \lambda_\sigma \mathcal{L}_\sigma +
    \lambda_{\boldsymbol c} \mathcal{L}_{\boldsymbol c} +
    \lambda_{\boldsymbol s} \mathcal{L}_{\boldsymbol s} +
    \lambda_\tau \mathcal{L}_\tau
\end{align}

Geometry completion $\mathcal{L}_\sigma$ is supervised in all of spacetime, while the other three loss terms $\mathcal{L}_{\boldsymbol c}$, $\mathcal{L}_{\boldsymbol s}$, $\mathcal{L}_\tau$ are applied in solid regions only.

\subsection{Inference}

After learning, we will be able to estimate a continuous representation of a point cloud %
from a video. For many applications, we need a sampling procedure to convert the continuous cloud into a discrete cloud. Depending on our choice of sampling technique, we can construct arbitrarily detailed point clouds at test time.

Since the target is unknown at test time, we sample query coordinates $(\boldsymbol p_q,t_q)$ uniformly at random within a 4D spacetime volume of interest. We generate discrete point clouds by filtering predictions according to solidity, only retaining a query point whenever the predicted occupancy is above some threshold, \ie $\hat y_\sigma \geq \sigma_T$.

For visualization purposes, we can also convert the predictions to scene meshes. The surface $\mathcal{S}$ of a mesh at time $t$ is implicitly defined as the zero-level set of the predicted occupancy $\hat\sigma$ relative to the threshold $\sigma_T$, \ie $\mathcal{S} = \{\, \boldsymbol x \in \mathbb{R}^3 \mid \hat y_\sigma(\boldsymbol x,t) = \sigma_T \,\}$, where $\sigma_T=0.5$.

After sampling a point cloud, or a mesh via the marching cubes algorithm, we colorize it by retrieving either the predicted color $\hat{\boldsymbol y_c}$, the semantic category $\hat{\boldsymbol y}_s$, or the tracking flag $\hat{y}_\tau$, associated with every coordinate.

\input{fig/res_greater}

\input{fig/res_carla}

\subsection{Implementation Details}

The feature encoder $\phi$ interleaves 4 self-attention layers with 3 down transition modules \cite{zhao2021point} to generate the featurized point cloud $\mathcal{Z}$ from $\mathcal{X}$. The continuous representation $f$, conditioned on $\mathcal{Z}$, accepts arbitrary 4D query coordinates $(\boldsymbol p_q,t_q)$ as input, applies Fourier encoding \cite{mildenhall2020nerf}, and interleaves 6 residual MLP blocks \cite{yu2020pixelnerf} with 2 cross-attention layers to produce $\hat{\boldsymbol y}$.

We feed in $T=12$ frames with $|\mathcal{X}|=14,336$ points in total, and train the model to predict the last $U=4$ frames, such that the first $T-U=8$ frames serve as an opportunity to aggregate and process spatiotemporal context.
More details can be found in the supplementary material.

%% file: fig/arch.tex
\begin{figure}[t]
\centering
\includegraphics[width=\linewidth]{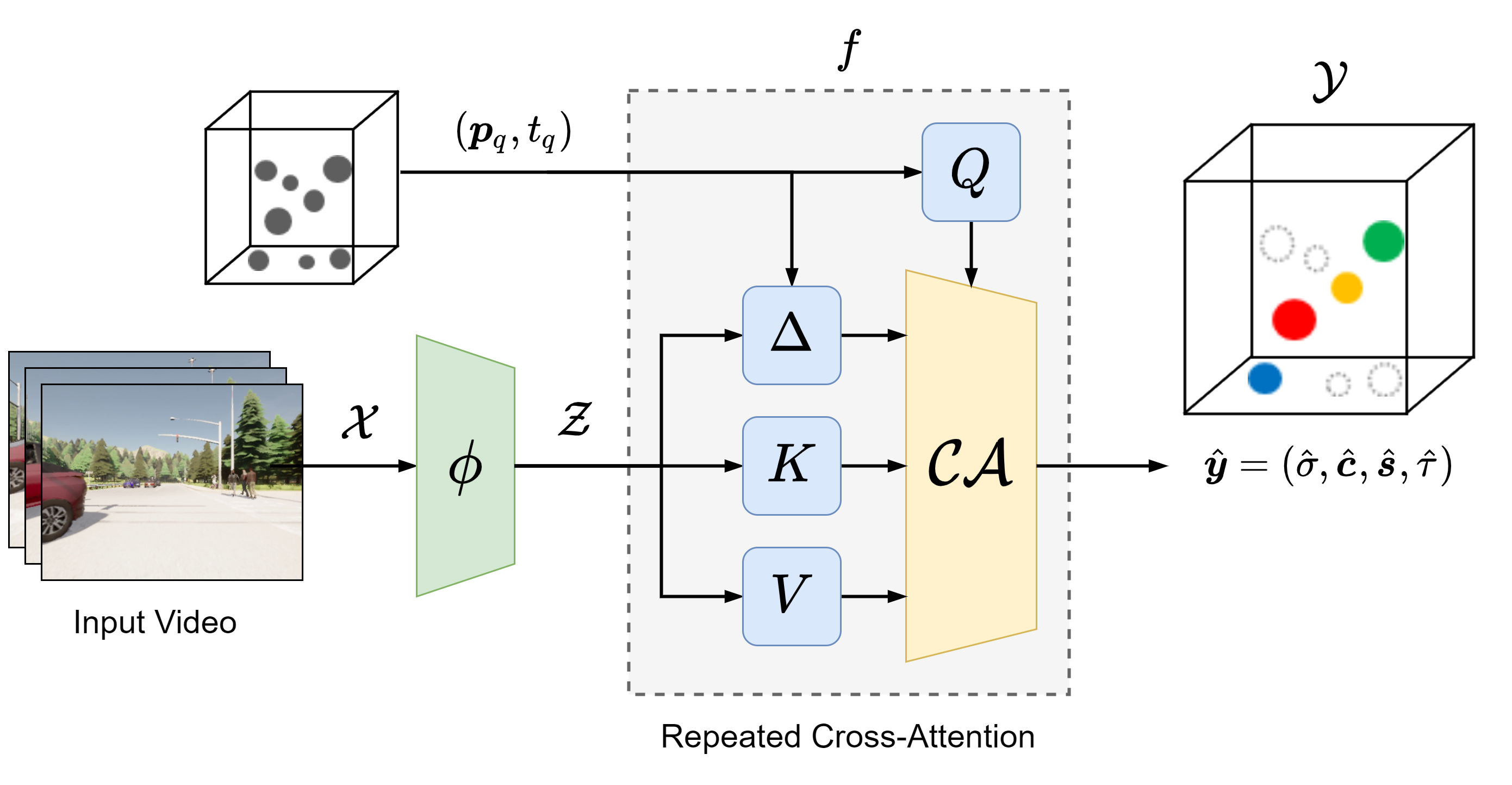}
\caption{
\textbf{Neural Architecture --}
The encoder $\phi$ is a point transformer that featurizes the input point cloud video $\mathcal{X}$ using self-attention to produce $\mathcal{Z}$. The implicit representation $f$, conditioned on $\mathcal{Z}$, incorporates cross-attention blocks to contextualize the query points $(\boldsymbol p_q,t_q)$ and create the desired output features $\hat{\boldsymbol y}$ for that location and time within the scene.
\vspace{-0.2cm}
}
\label{fig:arch}
\end{figure}

%% file: fig/keyidea.tex
\begin{figure}[t]
\centering
\includegraphics[width=0.95\linewidth]{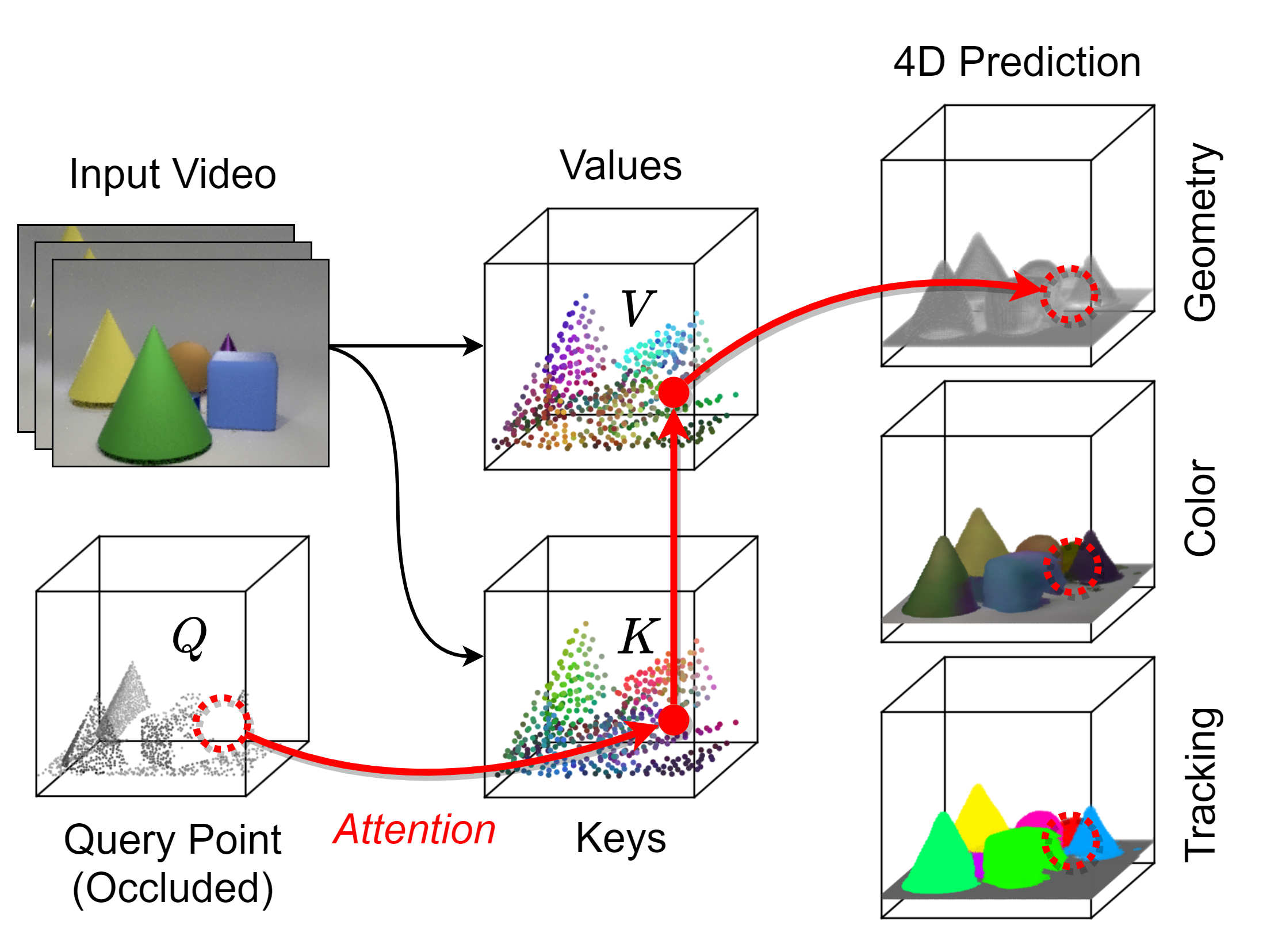}
\caption{
\textbf{Key Idea --}
Given a query point in 4D, the model learns to attend to keys and values extracted from the input video. When the query point corresponds to world coordinates that are occluded, the attention mechanism will learn to find the object when it was not yet occluded. When the world coordinates correspond to empty space, the model instead learns to predict a low occupancy $\hat y_\sigma \approx 0$. The network is flexible, and we can train the same model to produce point clouds for many different video tasks requiring object permanence.
\vspace{-0.2cm}
}
\label{fig:keyidea}
\end{figure}

%% file: fig/mvsuper.tex
\begin{figure}[t]
\centering
\includegraphics[width=\linewidth]{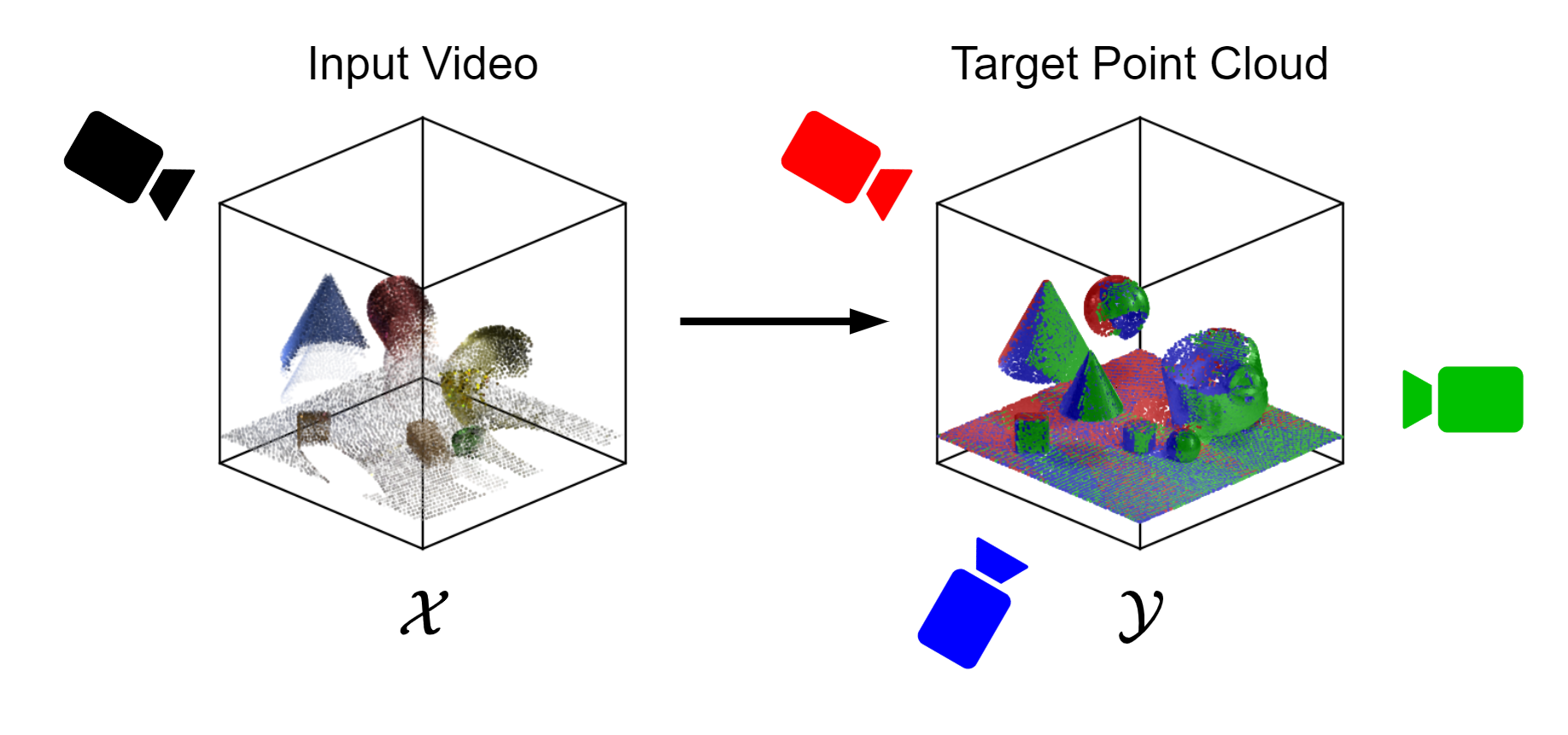}
\caption{
\textbf{Learning About Occlusions -- }
When an occlusion occurs in the input video, it is typically still visible from other viewpoints.
For example, the smaller cone becomes fully occluded in the last input frame, but is revealed again by the ground truth.
Our approach capitalizes on this natural clue to provide geometrically consistent multi-view self-supervision to the model, thus distilling the notion that objects persist across space and time.
\vspace{-0.2cm}
}
\label{fig:supervis}
\end{figure}

%% file: fig/res_greater.tex
\definecolor{mygreen}{rgb}{0.0, 0.8, 0.0}
\definecolor{myred}{rgb}{1.0, 0.0, 0.0}
\begin{figure*}
\centering
\includegraphics[width=0.9\linewidth]{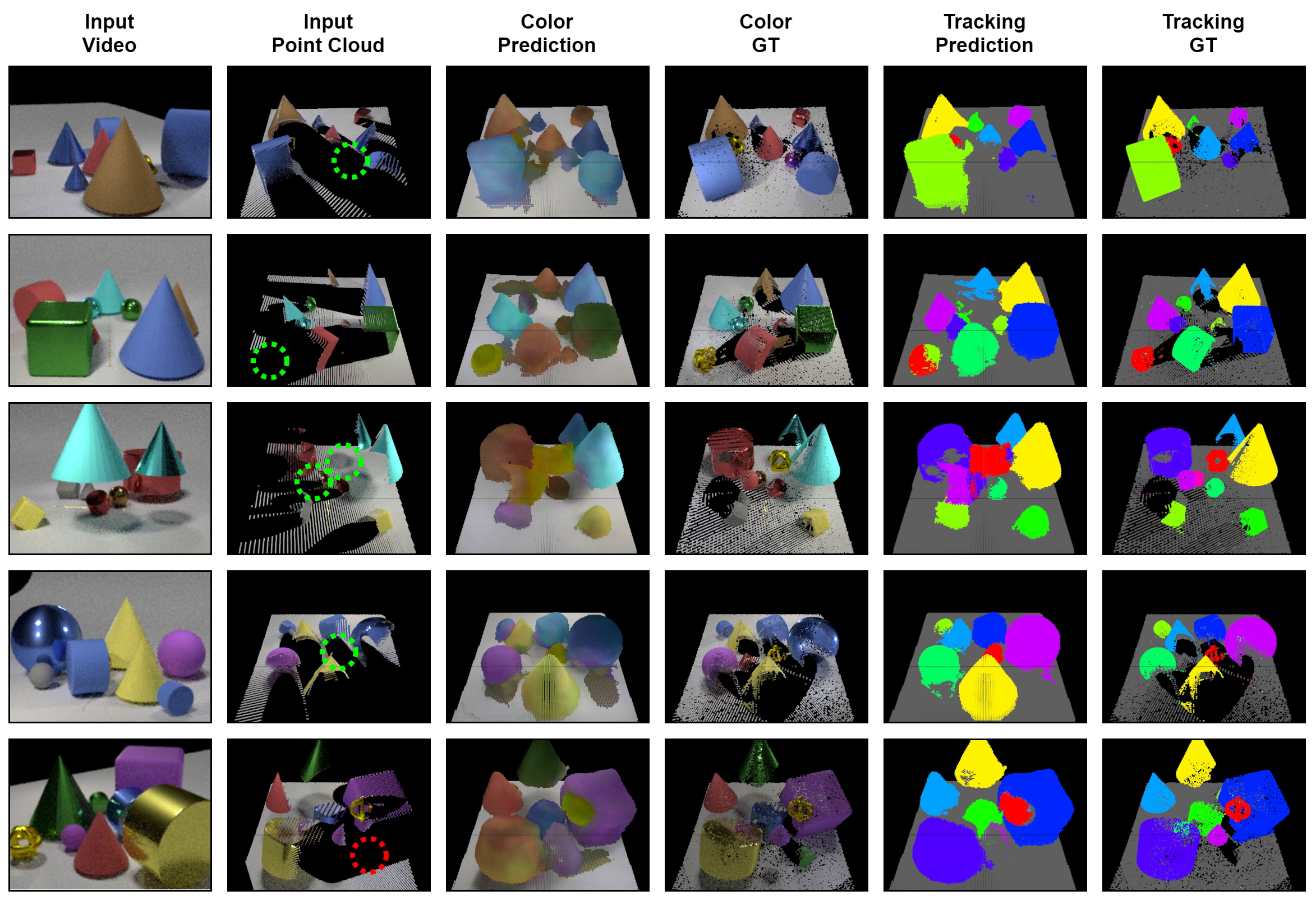}
\caption{
\textbf{Results for GREATER -- }
We show inputs, predictions, and ground truths. Our model receives color point clouds as input (second column), and we show the corresponding video frame in column one as reference. The third column represents \emph{both} the geometry reconstruction and color prediction tasks. We note how the model is able to (1) perform scene completion by filling in partially observed objects, \ie resolve amodal completion, and even (2) recover totally occluded objects, including when there are multiple occurring at once.
For \emph{total} occlusions, we circle the corresponding locations in the input for \textcolor{mygreen}{\textbf{true positives}} in green and \textcolor{myred}{\textbf{false negatives}} in red.
While we show only the last frames in this figure, the model predicts the scene at different time steps, capturing scene dynamics.
}
\label{fig:res_greater}
\end{figure*}

%% file: fig/res_carla.tex
\definecolor{mygreen}{rgb}{0.0, 0.8, 0.0}
\definecolor{myred}{rgb}{1.0, 0.0, 0.0}
\begin{figure*}
\centering
\includegraphics[width=\linewidth]{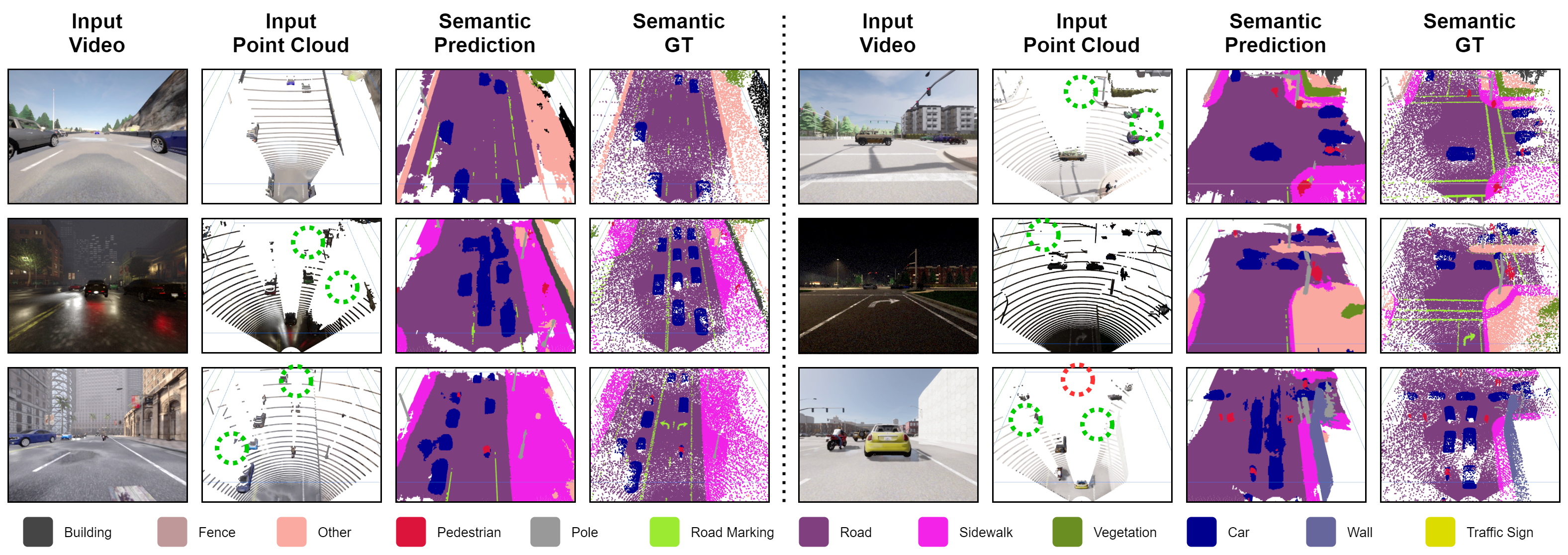}
\caption{
\textbf{Results for CARLA -- }
We show inputs, predictions and ground truths. Our model receives the point cloud video whose last frame is depicted in the second column, and predicts scene occupancy and semantic completion data for every sampled query point. Considering the limited input information, our model is capable of reconstructing the whole scene with high accuracy. Just as in Figure \ref{fig:res_greater}, all inputs and outputs are 4D meaning that they actually consist of multiple frames -- please see our \href{https://occlusions.cs.columbia.edu/}{webpage} for animated visualizations.
}
\label{fig:res_carla}
\end{figure*}

%% file: sec/4_dataset.tex
\section{Datasets}

In order to train and evaluate our model, particularly in terms of its ability to handle occlusions, we require multi-view RGB-D video from highly cluttered scenes.
To this end, we contribute two high-quality synthetic datasets, shown in \cref{fig:res_greater,fig:res_carla}. 
Brief descriptions %
are provided below, with further details in the supplementary material.

\subsection{GREATER}

We extend CATER \cite{girdhar2019cater} (which is in turn based on CLEVR \cite{johnson2017clevr}) mainly in order to increase the degree of occlusions, and call our proposed dataset GREATER.
Each scene in GREATER contains 8 to 12 cubes, cones, cylinders and spheres that move around, occluding one another in random ways. Partial and complete occlusions are happening constantly to the input view, which are only revealed by the target point clouds, allowing for effective learning and benchmarking of our model.
We capture 7,000 scenes lasting 12 seconds each, with data captured from 3 random views spaced at least 45\degree~apart horizontally, and a train/val/test split of 80\%/10\%/10\%.
On the GREATER dataset, we train our model to predict geometry, color, and tracking.

\subsection{CARLA}

While GREATER already exhibits many non-trivial scene configurations and movement patterns, it may be desirable to apply 4D video completion within more realistic environments as well.
Since object permanence is paramount for situational awareness in the context of driving and traffic scenarios, we employ the state-of-the-art driving simulator CARLA \cite{dosovitskiy2017carla} to generate a dataset of complex, dynamic road scenes. 
We sample 500 scenes lasting 100 seconds each, with data captured from 4 fixed views, and a train/val/test split of 80\%/8\%/12\%.
The scenes cover a wide variety of different towns, vehicles, pedestrians, traffic scenarios, and weather conditions.
On the CARLA dataset, we teach our model to perform geometric as well as semantic scene completion.

%% file: sec/5_experiments.tex
\section{Experiments}
\label{sec:experiments}

To display the generality of our framework, we test it on a variety of tasks across different datasets. Crucially, all the tasks can be trained end-to-end simultaneously in the same model. %
We set $(\lambda_\sigma, \lambda_{\boldsymbol c}, \lambda_{\boldsymbol s}, \lambda_\tau)=(1, 1, 0, 1)$ for GREATER and $(\lambda_\sigma, \lambda_{\boldsymbol c}, \lambda_{\boldsymbol s}, \lambda_\tau)=(1, 0, 0.6, 0)$ for CARLA.

\subsection{Evaluation Metrics}
We evaluate models using the Chamfer Distance (CD) metric between the predicted point cloud $\hat{\mathcal Y}$ and target point cloud $\mathcal Y$:
{\small
\begin{align}
    CD \left( \hat{\mathcal Y}, \mathcal Y \right)
    &= \frac{1} {|\hat{\mathcal Y}|} \sum_{i \in \hat{\mathcal Y}} \min_{j \in \mathcal Y}  |\boldsymbol p_i - \boldsymbol p_j|_2
    + \frac{1} {|\mathcal Y|} \sum_{j \in \mathcal Y} \min_{i \in \hat{\mathcal Y}} |\boldsymbol p_j - \boldsymbol p_i|_2 
\end{align}
}%
For geometry completion, we initially consider all points, but wish to specifically study occlusions as well. To that end, we filter all points by whether they belong to an occluded instance or not, which we can approximate by comparing different views with each other.
If the filtered output point cloud is empty (which typically corresponds to false negatives), we substitute the prediction for a single point at the center of the scene, as the CD would become undefined otherwise.

For instance tracking in GREATER, we track one object at a time, and merge the resulting predictions at test time. Concretely, we obtain multiple tracks by assigning the instance tag with the most confident score $\hat y_\tau$ to each point, but only if $\hat y_\tau \geq 0.5$.
Then, for every instance tag, we calculate the CD between only its corresponding predicted points and the ground truth object points, and subsequently average this value over all instances within a scene. We also report the average over occluded objects only.

For semantic segmentation in CARLA, we use a similar workflow as for tracking, but average over all categories instead of instances.
We study two important classes (pedestrians and vehicles) separately, which implies filtering both the predictions and targets by whether their ground truth semantic categories belong to those respective classes before reporting the CD values. Additionally, we filter for \emph{occluded} pedestrians and vehicles. In both cases, we average over all instances per scene such that every pedestrian or car is treated equally.

\input{tab/results}

\input{fig/time_occl_gr}

\subsection{Ablations and Baselines}
\paragraph{Ablations}
To show how various architectural choices affect our model's performance, we perform ablations to its four main components by: (1) removing local features from $f$; (2) removing the temporal dimension; (3) removing self-attention from $\phi$; (4) removing cross-attention from $f$; (5) combining (3) and (4).
Ablation (1) implies that instead of conditioning on $\mathcal Z$, we only pass a global embedding (that is the average of the features over all points in $\mathcal Z$) from $\phi$ to $f$. 
In (2), the model is no longer burdened with predicting a 4D dynamic representation consisting of multiple frames, and the task becomes 3D scene completion instead -- given a single frame, predict a single frame.
For (3), we replace self-attention layers in the point transformer with a simple point-wise linear projection.
For (4), since $f$ cannot attend to $\mathcal Z$ anymore, we feed in the nearest neighbor in $\mathcal Z$ of every query point to $f$ along with the query point itself.

\paragraph{Baselines}
For our primary scene geometry reconstruction task, we adapt Point Completion Network (PCN) \cite{yuan2018pcn} to our setup.
Additionally, we evaluate a `Copy input view' baseline, where the prediction is simply the input point cloud that the model sees, \ie the identity operation. Comparison with this baseline shows the benefits of our approach in revealing occlusions.
Finally, for tracking, we evaluate the baseline where the marked instance is propagated but remains stationary after the first frame, which is also the only time that the model sees its mask.

\subsection{Quantitative Results}
See \Cref{tab:greater,tab:carla}. %
Occlusion metrics (``Occ.'') are for objects that are more than 80\% occluded, as inferred from counting the number of points per instance that are visible from each view.
Our non-ablated model consistently outperforms most baselines and ablations with significant margins.
In particular, these results demonstrate that incorporating attention mechanisms is clearly beneficial for handling occlusions and performing spatiotemporal inpainting, suggesting that a robust notion of object permanence was successfully learned.

Although the ablation without time succeeds at reconstructing a 3D snapshot of the scene with relatively good quality, %
it is significantly worse at predicting occluded objects such as vehicles or pedestrians in CARLA, which is a critical aspect of interpreting traffic scenes.
Figure \ref{fig:time_occlusions} further demonstrates that temporal context is essential to understand scene dynamics.
Moreover, providing contextualized local features is also vital for the performance of the model.

\input{fig/attention}

\input{fig/uncertainty}

\subsection{Visualizations}

In this section, we visualize the inputs and predictions of our model, for both datasets.
Then, we focus our attention on how our model deals with two key challenges the task presents: occlusions and uncertainty.

Figures~\ref{fig:res_greater} and \ref{fig:res_carla} show our model predictions for the GREATER and CARLA datasets. In both cases, the model is capable of simultaneously performing geometry completion along with other prediction tasks such as visual reconstruction, instance tracking, or semantic segmentation. Note that geometry completion is a prerequisite to solving any other prediction task, since other tasks also require knowledge of objects that are {\em not visible} in the input frame.

Our model is capable of completing the scene with great detail, even when presented with a limited density of input points. Specifically, when trained on CARLA the model is capable of reconstructing---and predicting class information about---relatively small objects such as street poles (gray), pedestrians (red), or traffic signs (yellow). It does so for different temporal steps, and even when there are occlusions. We observe that the model trained on CARLA does sometimes struggle in hard cases that involve objects moving across long-term occlusions (which presents a limitation and opportunity for future work), but it usually generates an accurate, complete reconstruction in most other scenarios, especially when just amodal completion is involved.

In Figure~\ref{fig:time_occlusions} we visualize how our 4D model is capable of exploiting temporal context to track through occlusions, unlike the ``no time'' baseline.
We also show that the model is 4D in the sense that it can represent points for different time steps.

In Figure~\ref{fig:attention}, we visualize the attention of the model during occlusions in order to understand the mechanism it uses to resolve them. Specifically, we adapt attention rollout~\cite{abnar2020quantifying} to our architecture, and visualize the input points (across time) that contribute the most to the specific output class we want to analyze.
We show results for a pedestrian, two cars, and a motorcycle.
In all cases, the attention for the studied object is focused on its past trajectory, meaning that the model implicitly tracks objects through time in order to make predictions about their future.

Finally, the reader may have noticed some ``long cars'' in Figure~\ref{fig:res_carla}. These occur in places that remain occluded throughout the video, and Figure~\ref{fig:uncertainty} illustrates that our model appears to encode uncertainty with respect to the scene contents to some extent.
This allows us to control the degree of certainty we want to visualize the prediction at, and this parameter can optionally be tuned depending on the downstream application.

Please refer to the supplementary material for extra, non-cherry-picked visualizations that include more baselines.

%% file: tab/results.tex
\newcommand{\tablestyle}[2]{\setlength{\tabcolsep}{#1}\renewcommand{\arraystretch}{#2}\centering\footnotesize}
\newcolumntype{Y}{>{\centering\arraybackslash}X}

\begin{table}[t]
\centering
\footnotesize
\tablestyle{2.5pt}{1.05}
\begin{tabularx}{1.0\linewidth}{lYY|YY}
    \toprule
    \multicolumn{1}{c}{Method} & \multicolumn{2}{c}{Geometry} & \multicolumn{2}{c}{Tracking} \\
     & All & Occ. & Avg. Inst. & Occ. \\
    \midrule
    No local features & 0.78 & 0.73 & 4.50 & 4.86 \\ %
    No time & 0.26 & 0.49 & 1.59 & 4.25 \\ %
    No self-attention & 0.26 & 0.37 & 1.12 & 1.58 \\ %
    No cross-attention & 0.32 & 0.41 & 1.21 & 1.73 \\ %
    No attention & 0.40 & 0.48 & 1.42 & 2.00 \\ %
    \hline
    Copy input view & 0.48 & 1.92 & - & - \\
    Assume stationary & - & - & 2.34 & 3.43 \\
    PCN \cite{yuan2018pcn} & 0.59 & 0.97 & - & - \\
    \textbf{Ours} & \textbf{0.22} & \textbf{0.33} & \textbf{1.05} & \textbf{1.32} \\ %
    \bottomrule
\end{tabularx}
\vspace{-0.1cm}
\caption{\textbf{Results for GREATER -- geometry completion and instance tracking tasks}. We report the Chamfer Distance (lower is better). In addition to outperforming all ablations and baselines on both tasks, our model predicts occluded objects nearly as well as visible objects.}
\vspace{-0.3cm}
\label{tab:greater}
\end{table}

\setlength{\tabcolsep}{2pt}

\begin{table}
\centering
\footnotesize
\tablestyle{2.5pt}{1.05}
\begin{tabularx}{1.0\linewidth}{lY|YYYYY}
    \toprule
    \multicolumn{1}{c}{Method} & \multicolumn{1}{c}{Geometry} & \multicolumn{5}{c}{Semantic Segmentation}\\
     &
    All &
    Avg. Cls. &
    Ped. &
    Veh. &
    Occ. Ped. &
    Occ. Veh. \\
    \midrule
    No local features & 1.12 & 15.23 & 19.26 & 6.81 & 19.95 & 6.97 \\ %
    No time & 0.55 & 6.76 & 10.16 & 5.51 & 15.56 & 10.99 \\  %
    No self-attention & 0.50 & 6.60 & 5.19 & \textbf{3.29} & 5.98 & 4.73 \\ %
    No cross-attention & 0.73 & 7.11 & 7.57 & 4.22 & 11.33 & 7.55 \\ %
    No attention & 0.71 & 9.21 & 9.68 & 4.61 & 13.66 & 7.51 \\ %
    \hline
    Copy input view & 1.39 & - & - & - & - & - \\
    PCN \cite{yuan2018pcn} & 11.79 & - & - & - & - & - \\
    \textbf{Ours} & \textbf{0.47} & \textbf{5.82} & \textbf{4.76} & {3.42} & \textbf{5.71} & \textbf{4.14} \\ %
    \bottomrule
\end{tabularx}
\vspace{-0.1cm}
\caption{\textbf{Results for CARLA -- geometry completion and semantic completion tasks}. We report the Chamfer Distance (lower is better). Our model significantly outperforms almost all baselines and ablations, especially for occluded pedestrians (``Ped.'') and vehicles (``Veh.'').}
\vspace{-0.4cm}
\label{tab:carla}
\end{table}

%% file: fig/time_occl_gr.tex
\begin{figure}[t]
\centering
\includegraphics[width=0.9\linewidth]{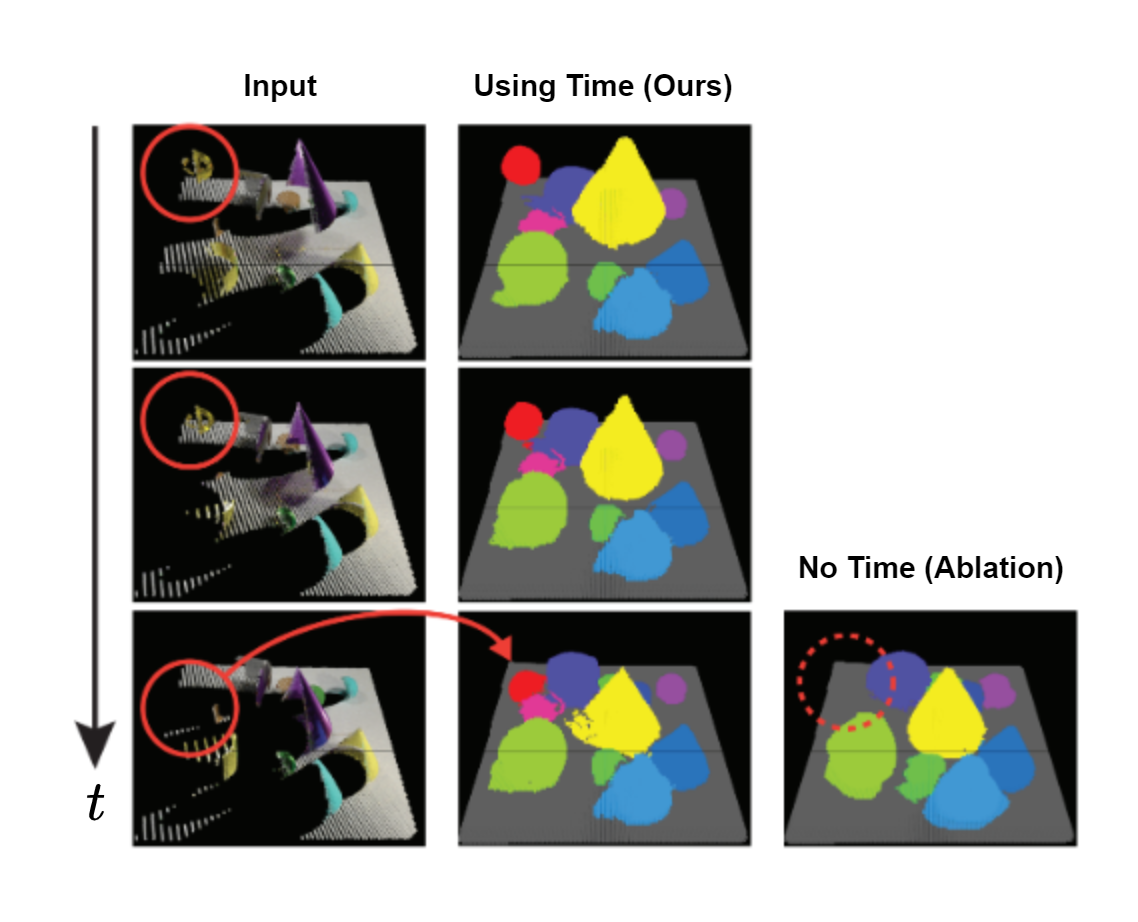}
\vspace{-0.4cm}
\caption{
\textbf{Importance of Time --}
In this figure, we show a video where an occlusion takes place (see yellow orb, highlighted by the red circle), and tracking predictions of our model. Because our model takes video as input, it is capable of tracking the orb through the occlusion. The single-frame baseline (``no time'') cannot recover the occluded orb, as it does not have access to temporal context. This example shows that (1) temporal context is important, and (2) our model learns to use it.
\vspace{-0.2cm}
}
\label{fig:time_occlusions}
\end{figure}

%% file: fig/attention.tex
\definecolor{mypedclr}{rgb}{1.0, 0.5, 0.0}
\definecolor{myattnclr}{rgb}{0.0, 0.8, 0.4}
\begin{figure}[t]
\centering
\includegraphics[width=\linewidth]{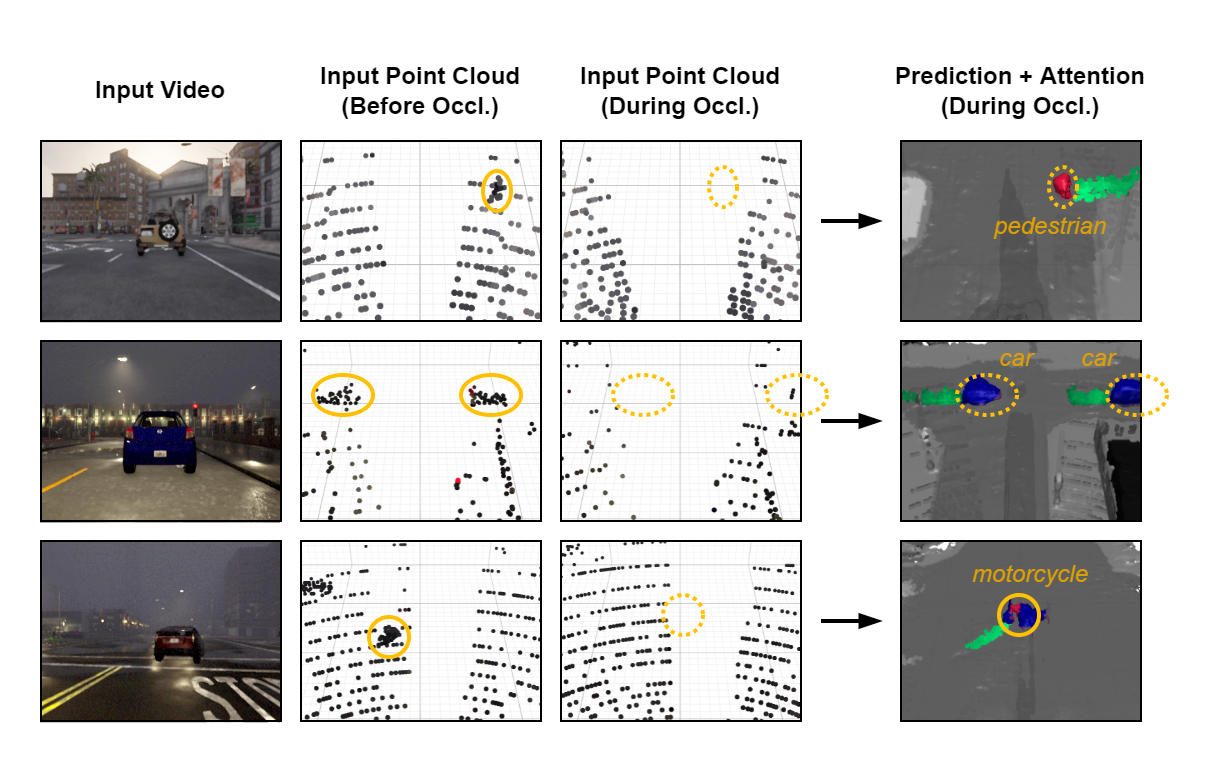}
\vspace{-0.6cm}
\caption{
\textbf{Visualizing Attention -- }
Why did the model predict the \textcolor{mypedclr}{\textbf{occluded object(s)}}? By backtracking neuron activations through all cross-attention and self-attention layers, we see that a mechanism of temporal correspondence emerges. In this example, the \textcolor{myattnclr}{\textbf{attention weights}} highlight input points that represent the trajectory of the object(s) over time, which suggests that our model implicitly learns to track them in order to succeed at 4D scene completion.
\vspace{-0.2cm}
}
\label{fig:attention}
\end{figure}

%% file: fig/uncertainty.tex
\begin{figure}[t]
\centering
\includegraphics[width=\linewidth]{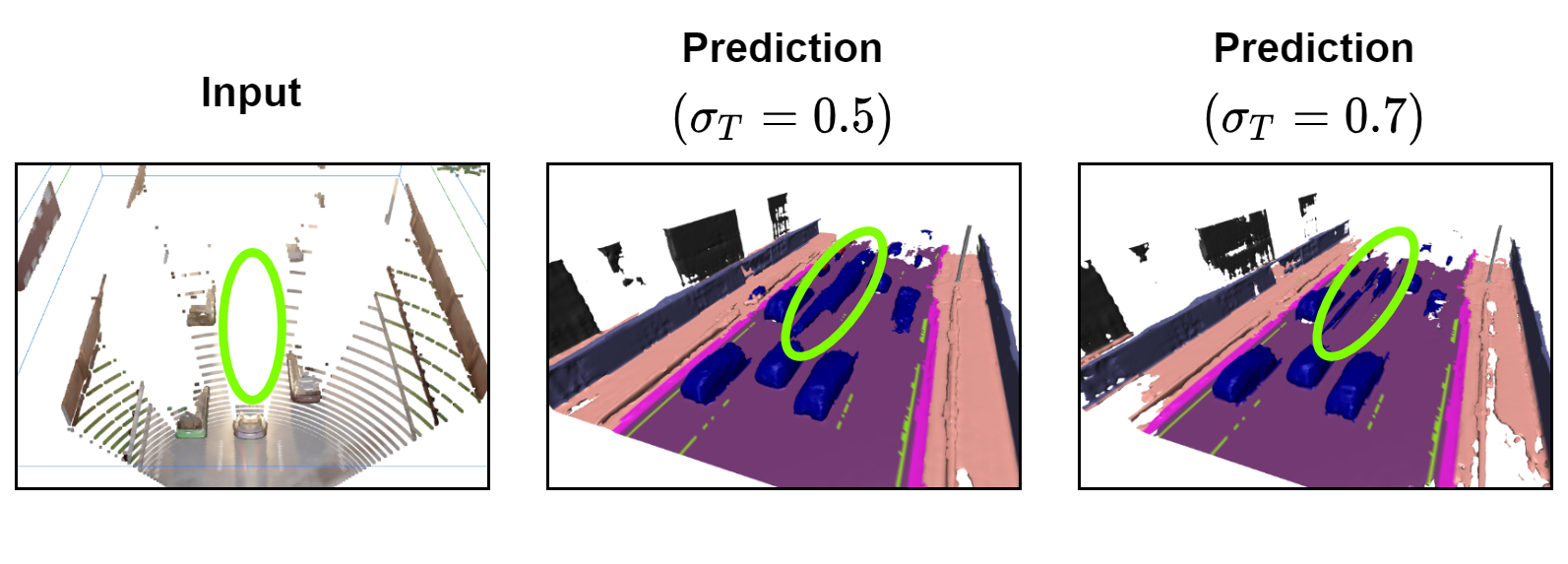}
\vspace{-0.6cm}
\caption{
\textbf{Interpreting Uncertainty -- }
Because it is often hard to perceive what is ahead of the car in front of you, some regions of the input can remain unseen in the LiDAR point cloud (shown on the left) throughout the entire input video, causing artefacts such as a `long car' to crop up. By varying the occupancy threshold $\sigma_T$ of the implicit surface, we can control the degree of certainty at which to visualize the prediction. For example, in the above example the model is found to be less certain about the presence of the `long car' relative to the rest of the scene (which remains reconstructed accurately), as indicated by the artefact's disappearance with increasing $\sigma_T$.
\vspace{-0.2cm}
}
\label{fig:uncertainty}
\end{figure}

%% file: sec/6_conclusions.tex
\section{Discussion}
We introduce the task of 4D dynamic scene completion, along with two datasets for understanding occlusions, and showcase a continuous representation that incorporates cross-attention as an initial attempt toward solving this challenge. We believe these these techniques and benchmarks will be useful in the context of scene completion, spatiotemporal inpainting, and object permanence.

\textbf{Acknowledgements:} This research is based on work supported by Toyota Research Institute, the NSF CAREER Award \#2046910, and the DARPA MCS program under Federal Agreement No. N660011924032. DS is supported by the Microsoft PhD Fellowship. The views and conclusions contained herein are those of the authors and should not be interpreted as necessarily representing the official policies, either expressed or implied, of the sponsors.

%% file: sec/X_supplementary.tex
\appendix

\setcounter{page}{1}

\twocolumn[
\centering
\Large
\textbf{Revealing Occlusions with 4D Neural Fields} \\
\vspace{0.5em}Supplementary Material \\
\vspace{1.0em}
] %
\appendix

\input{fig/detailarch}

\section{Implementation Details}

\subsection{Architecture and Hyperparameters}

We illustrate the full model architecture in Figure \ref{fig:detailarch}.
The input point cloud video $\mathcal{X}$ is subsampled using a mixture of random point sampling and farthest point sampling to contain precisely 14,336 points with 7 to 8 features each.
After the first MLP, the embedding size per point is 36, which becomes double at every down transition step, such that $\mathcal{Z}$ has 288 features per embedding. The encoder $\phi$ also averages all features into a 128-dimensional global embedding, which is also passed to $f$ alongside the featurized point cloud $\mathcal{Z}$. The latent size of embeddings within $f$ is 416 across all its components.

The self-attention block causes every point to attend to its 16 nearest neighbors \cite{zhao2021point}.
The down transition block selects one third of all incoming points by means of farthest point sampling, and subsequently performs channel-wise max-pooling from its 12 nearest neighbors.
The residual MLP block in $f$ initially distills a weighted linear interpolation (based on Euclidean distance) of the 8 nearest neighbors in $\mathcal{Z}$ around the query point as a starting point, after which cross-attention has the chance to apply a \emph{learned} interpolation of embeddings instead.
The cross-attention block causes every query point to attend to its 14 nearest neighbors in the featurized input points $\mathcal{Z}$, \ie $|\mathcal{M}(i)|=14,\,\forall\,i$.

Using the AdamW optimizer \cite{loshchilov2017decoupled}, we train two separate models (one per dataset) over 20 epochs for GREATER, and 40 epochs for CARLA. Our model takes between 18 and 55 hours to train on two RTX A6000 GPUs, and dense inference across the entire spacetime cube takes roughly one minute. The initial learning rate is $0.001$, but this drops by a factor $2.5$ at progress rates of 40\%, 60\%, and 80\%.

\subsection{Point Sampling}

During training, within every frame we sample 7,168 solid query points ($\sigma=1$) and 10,752 free space (air) query points ($\sigma=0$). The air points are uniformly randomly sampled within the output box of interest, except if they are within a distance of $2\epsilon=0.2$ within any target point in $\mathcal Y$. The solid points are selected as a random subset of the target point cloud, but we add a small random spatial offset to every solid query point that itself is uniformly sampled within a spherical ball of radius $\epsilon=0.1$. This roughly corresponds to the spacing between target points on the objects and the floor in the dataset, encouraging the model to learn to "fill in the gap" between those points.

In the case of CARLA, to ensure that we maintain an effective learning signal, we describe several tricks that help guide supervision toward areas where it is deemed more important.
In addition to random sampling from the target point cloud, 14\% of solid points are explicitly sampled in dynamic regions of the scene, \ie moving points that did not exist in a randomly selected other frame. The converse is also done for air points (\ie regions that are now missing, but were present in another frame).
At least 7\% of sampled solid points focus exclusively on vehicles and pedestrians.
We also oversample occluded vehicles and pedestrians (for up to 7\% of all sampled solid points), by counting and comparing the number of points of every object seen by every view. Moreover, we ensure that objects that were never seen in the first place (\eg pedestrians who remain behind a building or wall throughout the entire video) are not oversampled, to avoid confusing the model.
Lastly, for class balancing, 14\% of sampled solid points treat all semantic categories equally, \ie we sample the same number of points from every class that is present in the scene.

For a fair and correct evaluation, there are no sampling tricks at test time, \ie we apply uniform sampling within the cuboid of interest in a way that is agnostic of the ground truth. For GREATER, we sample $N=2^{19}$ points per frame per scene, while for CARLA, $N=2^{21}$.

\subsection{Evaluation Metrics}

The model is evaluated by the Chamfer Distance (CD) between the sampled prediction and the ground truth point cloud. Sometimes, the model fails to predict an object when it is fully occluded (\ie a false negative),
which may cause the output point cloud (filtered by the desired category and occlusion rate) to consist of zero points. Rare or "tiny" classes with a low number of target points per scene, \eg traffic sign, may face similar issues. In that case, the CD would normally be undefined, but in order to ensure that the average metric
accounts for this and is still affected,
we substitute the prediction with a single point in the center of the scene: $(0, 0, 0)$ for GREATER, or $(20, 0, 0)$ for CARLA. %
Moreover, unless explicitly noted otherwise, we set $\sigma_T=0.4$ for numerical evaluations (\ie Tables \ref{tab:greater} and \ref{tab:carla}) to reduce the likelihood of false negatives, and $\sigma_T=0.5$ for visualization purposes (\ie Figures \ref{fig:res_greater} and \ref{fig:res_carla}).

\input{fig/dset_mv_supp}

\section{Dataset Description}

Both GREATER and CARLA are posed multiview RGB-D video datasets, with added instance segmentation and semantic segmentation annotations respectively. The camera views are illustrated in Figures \ref{fig:gr_mv} and \ref{fig:ca_mv}.

\subsection{GREATER}

All videos are recorded with a virtual RGB-D camera with known intrinsics and extrinsics. The dataset is generated at 24 frames per second (FPS), and objects move and/or rotate in synchronized cycles that repeat every 32 to 42 frames.
The model's data loader subsamples temporally and uses 8 FPS, such that along with $T=12$, roughly one full cycle is covered per clip.
All objects are precisely twice as large as compared to CATER \cite{girdhar2019cater}. For every scene, the number of objects is selected uniformly at random between 8 and 12 (inclusive). The cameras are also chosen randomly per scene, but henceforth remain static (\ie never move over time) over the duration of a single scene. With the spatially vertical axis denoted $z$, the 3D bounding box within which both the input and predictions happen is $x \in [-5, 5]$, $y \in [-5, 5]$, $z \in [-1, 5]$.

\subsection{CARLA}

All videos are recorded with a pair of sensors with known intrinsics and extrinsics: one RGB camera, and one LiDAR sensor. The point cloud data generated by the latter sensor does not contain color information, so we use the RGB images to colorize the points. While the dataset is generated at 10 FPS, the model's data loader subsamples temporally and uses 5 FPS.
Both the LiDAR and camera horizontal fields of view are 120 degrees. However, the LiDAR's spherical geometry is different from a camera's projective geometry. implying that the LiDAR data is not directly aligned with the camera intrinsics. Therefore, in order to obtain a colorized point cloud per frame, we first project all LiDAR points onto the image and then map the pixel's RGB values it was assigned to back to the 3D point. If a LiDAR point falls outside of the camera's field of view, we mark it with a generic "color unavailable" constant, \ie $(-1, -1, -1)$.

For every input clip passed to the model, since the vehicle pose over time is known, we correct all point clouds to a common reference frame. This reference frame is chosen to be the last input (and output) frame, such that the pair of sensors mounted to the ego vehicle is always at $(0, 0, 1)$ at time $t=T-1$. With the forward axis denoted $x$, the sideways axis denoted $y$, and the vertical axis $z$, the input bounding box (\ie containing all observed points) is $x \in [-14, 50]$, $y \in [-20, 20]$, $z \in [-1, 10]$, and the output bounding box (\ie containing all predicted and ground truth points) is $x \in [0, 40]$, $y \in [-16, 16]$, $z \in [-1, 6.4]$ (in meters).

\subsection{Clip Sampling}

During training, for GREATER, we sample clips uniformly randomly. For CARLA however, most clips are relatively uninteresting, and we encourage learning about occlusions by performing biased clip sampling. Specifically, we construct a subset of starting frame indices where we know (as derived from semantic information over time in the LiDAR point clouds provided by the simulator) that occlusions are more likely to happen, and return a clip from this pool 40\% of the time.
To preemptively avoid overfitting, the data loader will never return the exact same clip twice over the entire duration of training.

During testing and evaluation, we deterministically sample a single clip within every video that has the most occlusions happening at once, counted over the number of objects. This implies that the test set for CARLA is significantly more challenging than the average driving situation (\ie as compared to if we were to sample clips uniformly at random).

\input{fig/supp_res}

\section{Qualitative Results}

Figures \ref{fig:supp_res_gr} and \ref{fig:supp_res_ca} showcase \emph{non-cherry-picked} visualizations made by our model as well as two important baselines. In terms of predictions, we depict our non-ablated 4D dynamic scene completion model, the ablation without time (which essentially becomes a 3D scene completion model without dynamics), and finally the PCN \cite{yuan2018pcn} baseline. Note that even though our adaptation of PCN can see temporal context and is given an advantage by borrowing features from the ground truth point cloud, the scenes in our dataset appear to be too complicated for PCN to learn effectively, especially in the case of CARLA.

Finally, please see our webpage at \url{occlusions.cs.columbia.edu} for more visualizations, as well as links to our datasets, source code, and models.

%% file: fig/detailarch.tex
\begin{figure*}[t]
\begin{center}
\includegraphics[width=0.9\linewidth]{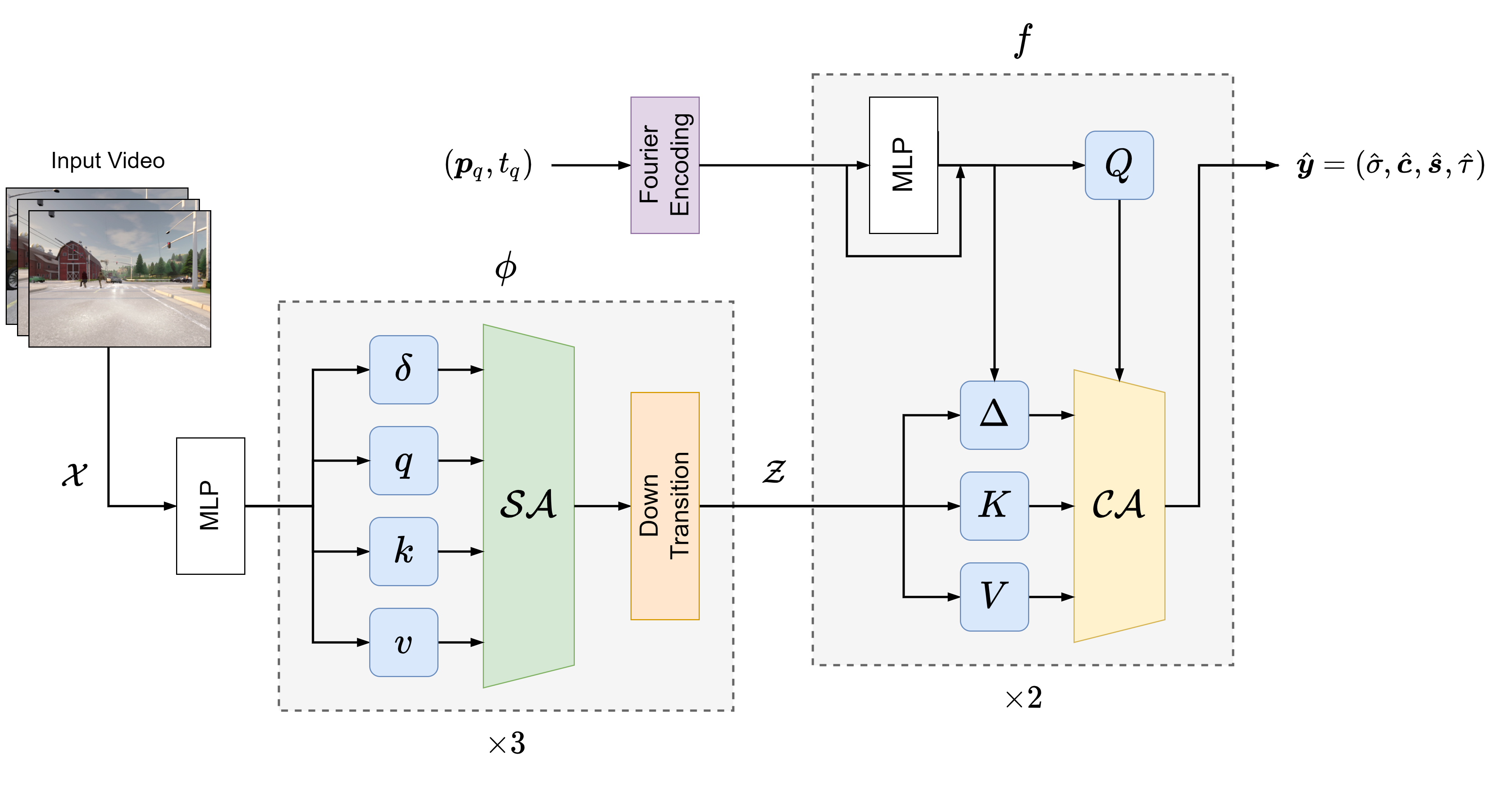}
\end{center}
\caption{
\textbf{Detailed Architecture -- }
We show the feature encoder $\phi$ and implicit representation $f$, with self-attention $\mathcal{SA}$ and cross-attention $\mathcal{CA}$ operations respectively. $\phi$ is a point transformer with four self-attention layers in total (the one after the last down transition is not shown),
although the outputs of the last two down self-attention blocks are combined to form multi-scale features for $\mathcal Z$. The exact operation of the down transition modules is described in \cite{zhao2021point}.
For the query points $(\boldsymbol p_q, t_q)$, we adopt the same Fourier encoding as \cite{mildenhall2020nerf}. All MLP blocks consist of two linear layers with a ReLU non-linearity in-between. The MLP blocks in the implicit representation $f$ are residual, similar to \cite{yu2020pixelnerf}.
}
\label{fig:detailarch}
\end{figure*}

%% file: fig/dset_mv_supp.tex
\begin{figure}[t]
\centering
\includegraphics[width=\linewidth]{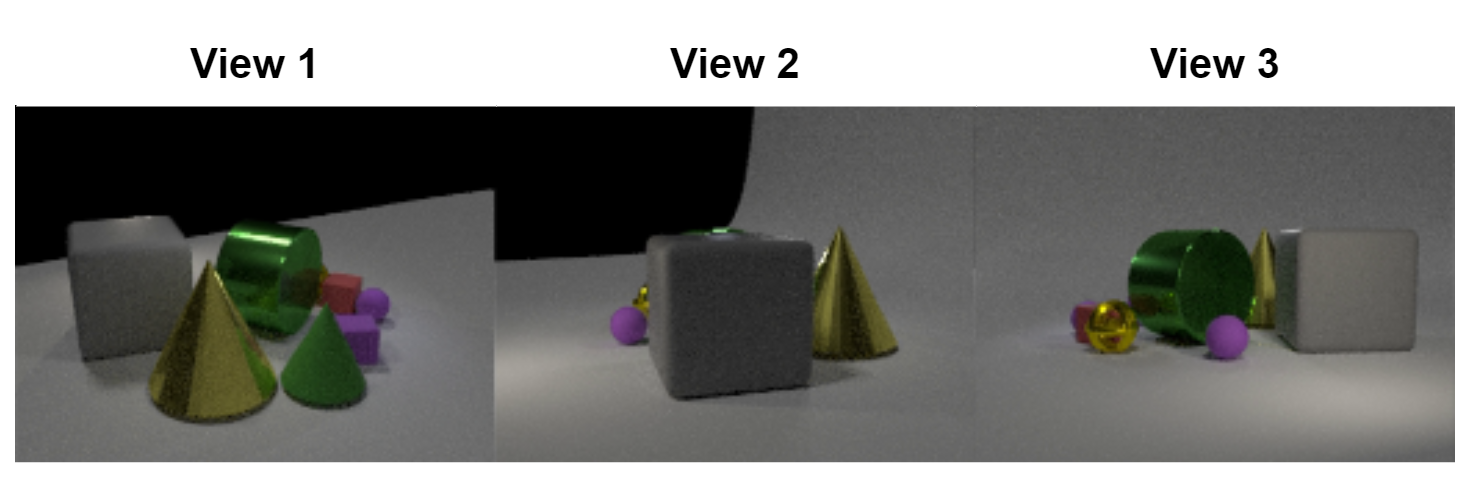}
\caption{
\textbf{GREATER Dataset Views -- }
Every GREATER scene has three non-moving RGB-D cameras at uniformly random azimuth angles, with the condition that all three are spaced by at least 45\degree\ away from each other. A random view is always selected to serve as input view, such that the other two views (along with the input view itself) serve as supervision during training.
}
\label{fig:gr_mv}
\end{figure}

\begin{figure}[t]
\centering
\includegraphics[width=0.9\linewidth]{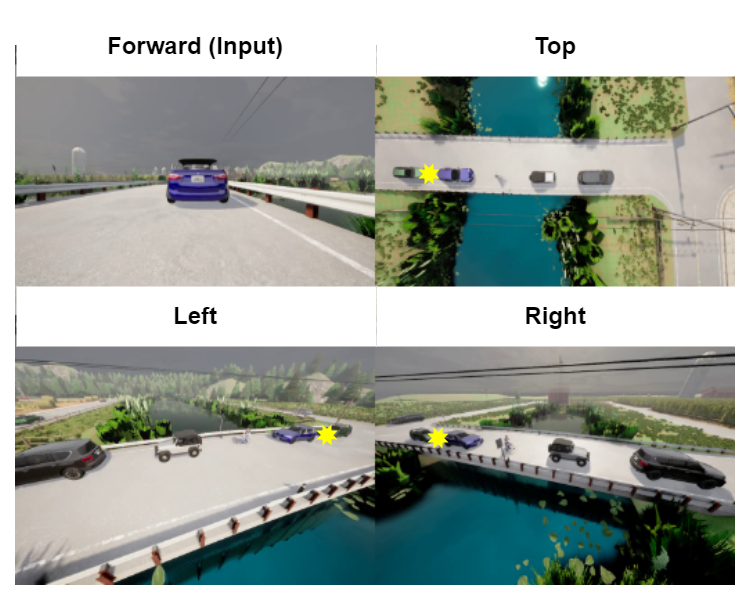}
\caption{
\textbf{CARLA Dataset Views -- }
Every CARLA scene has one RGB and LiDAR sensor attached to the front of the ego vehicle, which always records the input video. Three other supervisory views (in which the forward sensor position is marked with a yellow asterisk) operate only at training time.
}
\label{fig:ca_mv}
\end{figure}

%% file: fig/supp_res.tex
\begin{figure*}
\centering
\includegraphics[width=\linewidth]{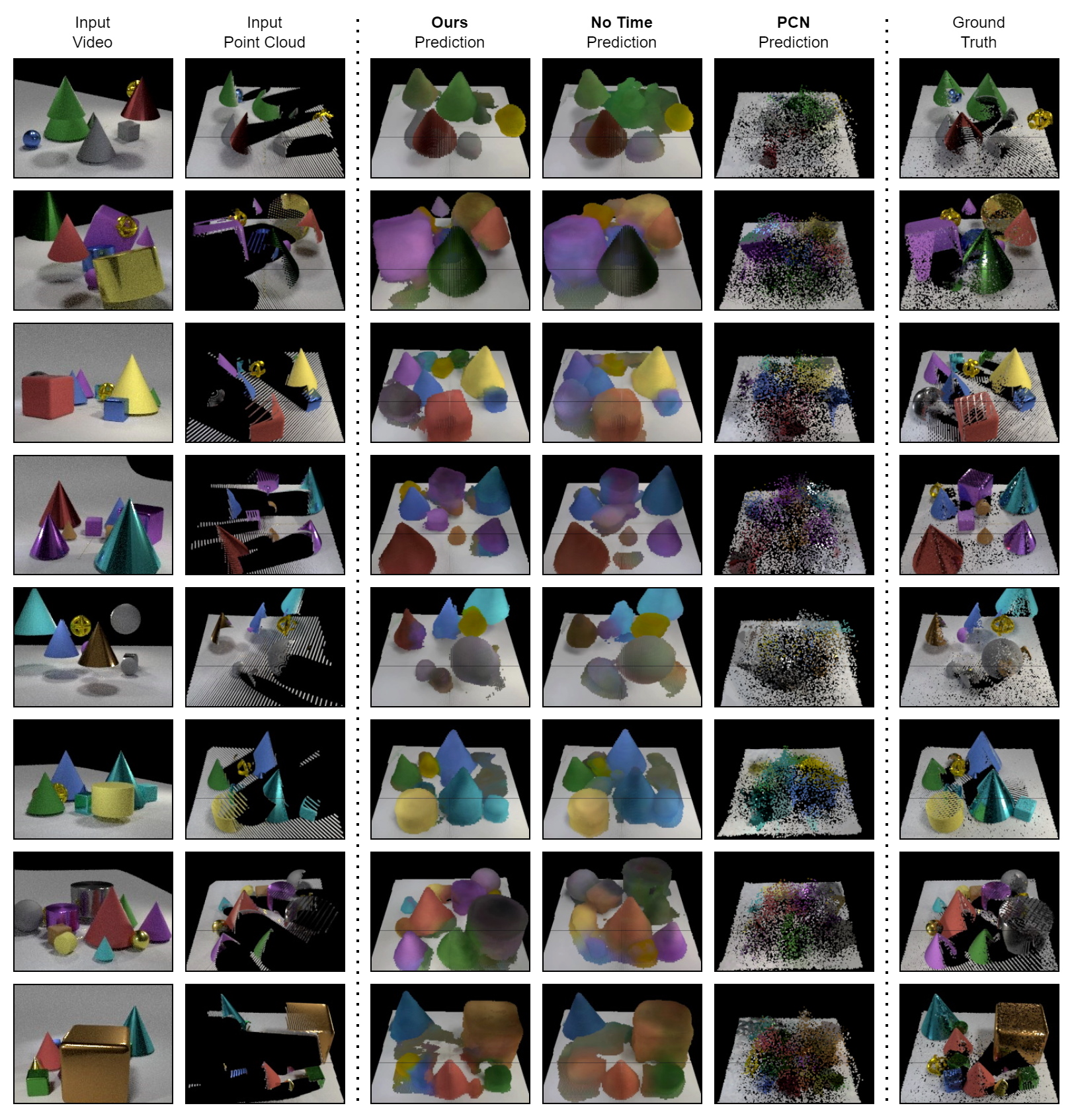}
\caption{
\textbf{Non-cherry-picked results for GREATER -- }
We show the last input frame, last output frames, and corresponding ground truth point cloud. Note that PCN does not have a mechanism built-in for predicting features (in this case, RGB color) associated with points, so we \emph{copy colors from the nearest ground truth point} in order to boost the output legibility.
Our model qualitatively has the best understanding of object permanence, and visually outperforms the baselines.
}
\label{fig:supp_res_gr}
\end{figure*}

\begin{figure*}
\centering
\includegraphics[width=\linewidth]{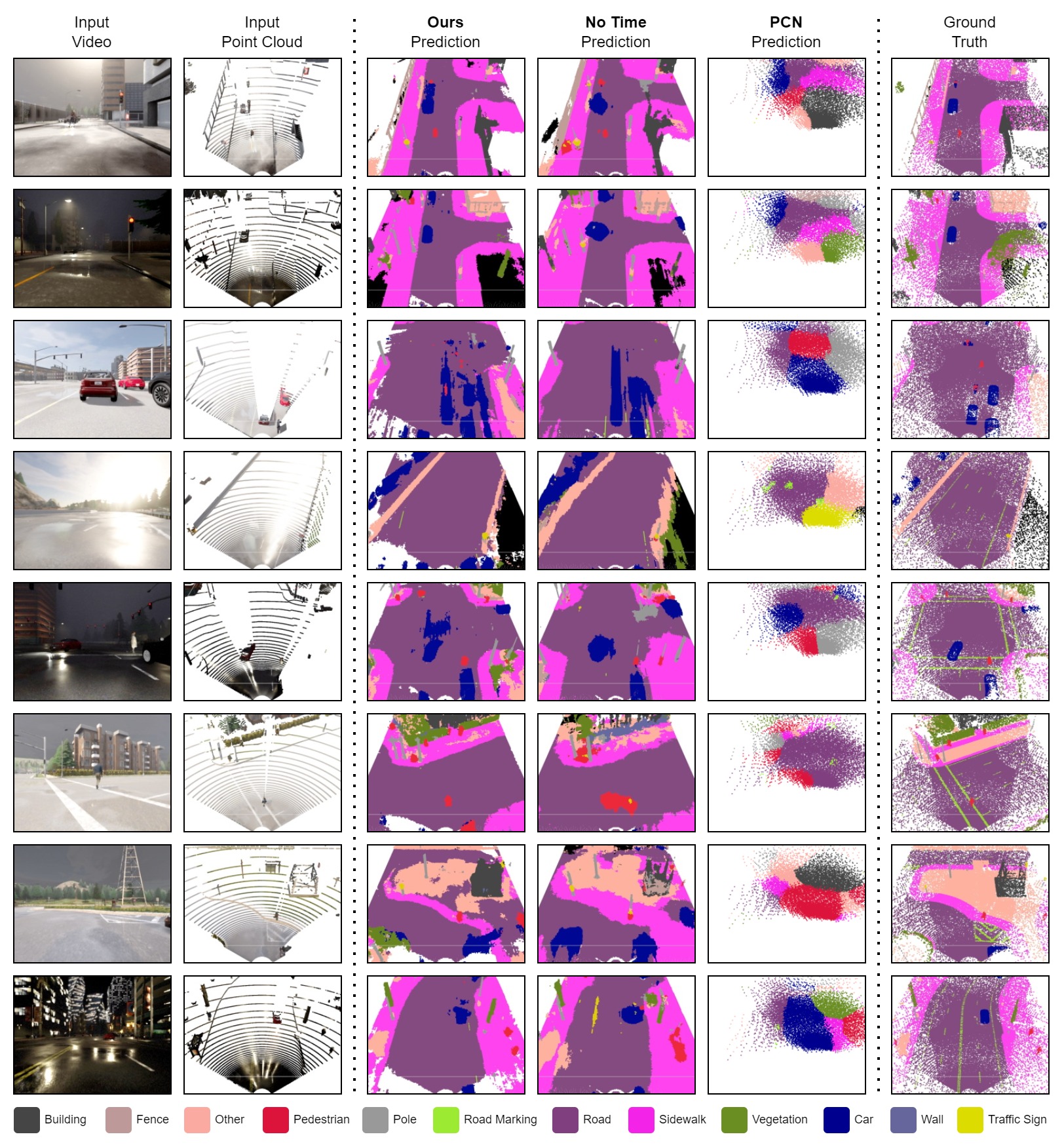}
\caption{
\textbf{Non-cherry-picked results for CARLA -- }
We show the last input frame, last output frames, and corresponding ground truth point cloud. Note that PCN does not have a mechanism built-in for predicting features (in this case, semantic category) associated with points, so we \emph{copy category information from the nearest ground truth point} in order to boost the output legibility.
Our model qualitatively has the best understanding of object permanence, and visually outperforms the baselines.
}
\label{fig:supp_res_ca}
\end{figure*}